\documentclass[conference]{IEEEtran}
\usepackage[utf8]{inputenc}
\usepackage[english]{babel}
\usepackage[T1]{fontenc}
\usepackage{amsmath}
\usepackage{amsthm}
\usepackage{amsmath}
\usepackage{mathtools}

\usepackage{color,colortbl}
\usepackage{flushend}

\usepackage{amsfonts}
\usepackage{amssymb}
\usepackage{graphicx}
\usepackage{caption}
\usepackage{subcaption}
\usepackage{array}
\usepackage{tikz}
\usepackage{placeins}
\usepackage{stmaryrd}
\usepackage{textcomp}
\usepackage{bbm}
\usepackage{bm}
\usepackage{algpseudocode}
\usepackage{algorithm}
\usepackage{multirow}

\usepackage[colorlinks = true,
            citecolor = blue,
            bookmarks=false
            ]{hyperref}

\newcommand{\ie}{\emph{i.e.}, }
\newcommand{\eg}{\emph{e.g.}, }
\newcommand{\R}{\mathbbm{R}}

\DeclareMathOperator*{\argmin}{arg\,min}

\renewcommand*\Call[2]{\textproc{#1}(#2)}

\makeatletter
\providecommand*{\diff}%
{\@ifnextchar^{\DIfF}{\DIfF^{}}}
\def\DIfF^#1{%
\mathop{\mathrm{\mathstrut d}}%
\nolimits^{#1}\gobblespace}
\def\gobblespace{%
\futurelet\diffarg\opspace}
\def\opspace{%
\let\DiffSpace\!%
\ifx\diffarg(%
\let\DiffSpace\relax
\else
\ifx\diffarg[%
\let\DiffSpace\relax
\else
\ifx\diffarg\{%
\let\DiffSpace\relax
\fi\fi\fi\DiffSpace}

\usepackage{xspace}
\newcommand{\D}{$\mathcal{D}$\xspace}
\newcommand{\G}{$\mathcal{G}$\xspace}

\title{MD-GAN: Multi-Discriminator Generative Adversarial Networks for Distributed Datasets}
\author{\IEEEauthorblockN{ Corentin Hardy}
\IEEEauthorblockA{\textit{Technicolor}, \textit{Inria}\\
Rennes, France}
\and
\IEEEauthorblockN{Erwan {Le Merrer}}
\IEEEauthorblockA{\textit{Inria}\\
Rennes, France}
\and
\IEEEauthorblockN{Bruno Sericola}
\IEEEauthorblockA{\textit{Inria} \\
Rennes, France}
}
\date{June 2018}

\begin{document}

\maketitle

\begin{abstract}

A recent technical breakthrough in the domain of machine learning is the discovery and the multiple applications of Generative Adversarial Networks (GANs). Those generative models are computationally demanding, as a GAN is composed of two deep neural networks, and because it trains on large datasets. A GAN is generally trained on a single server.

In this paper, we address the problem of distributing GANs so that they are able to train over datasets that are spread on multiple workers. MD-GAN is exposed as the first solution for this problem: we propose a novel learning procedure for GANs so that they fit this distributed setup. 
We then compare the performance of MD-GAN to an adapted version of \textit{Federated Learning} to GANs, using the MNIST and CIFAR10 datasets. MD-GAN exhibits a reduction by a factor of two of the learning complexity on each worker node, while providing better performances than federated learning on both datasets. We finally discuss the practical implications of distributing GANs.

\end{abstract}

\section{Introduction}

Generative Adversarial Networks (GANs for short) are \textit{generative} models, meaning that they are used to generate new realistic data from the probability distribution of the data in a given 
dataset.  Those have been introduced by Goodfellow \textit{et al} in seminal work \cite{gan}.  Applications are for instance to generate pictures from text descriptions \cite{ganText2Img}, to generate video from still images \cite{ganImg2Vid}, to increase resolution of images \cite{resolution}, or to edit them \cite{editImg}. Application to the chess game \cite{DBLP:journals/corr/ChidambaramQ17} or to anomaly detection \cite{anomaly} were also proposed, which highlights the growing and cross-domain interest from the machine learning research community towards GANs.

A GAN is a machine learning model, and more specifically a certain type of deep neural networks. As for all other deep neural networks, GANs require a large training dataset in order to fit the target application. Nowadays, the norm is then for service providers to collect large amounts of data (user data, application-specific data) into a central location such as their datacenter; the learning phase is taking place in those premises.
The image super-resolution application \cite{resolution} for instance leverages $350,000$ images from the ImageNet dataset; this application is representative of new advances: it provides state of the art results in its domain (measured in terms of quality of image reconstruction in that example); yet the question of computational efficiency or parallelism is left aside to futureworks.

The case was made recently for geo-distributed machine learning methods, where the data acquired at several datacenters stay in place \cite{gaia,geo}, as the considered data volumes would make it impossible to meet timing requirements in case of data centralization. Machine learning algorithms are thus to be adapted to that setup.
Some recent works consider multiple generators and discriminators with the goal to improve GAN convergence \cite{MGAN,mGGAN}; yet they do not aim at operating over spread datasets.
The \textit{Parameter Server} paradigm \cite{ParameterServer_OSDI2014} is the prominent way of distributing the computation of classic (\ie non-GAN) neural networks: workers compute the neural network operations on their data share, and communicate the updates (gradients) to a central server named the parameter server. This framework is also the one leveraged for geo-distributed machine learning \cite{gaia}.

In this paper we propose MD-GAN, a novel method to train a GAN in a distributed fashion, that is to say over the data of a set of participating workers (\eg datacenters connected through WAN \cite{gaia}, or devices at the edge of the Internet \cite{nca1}). 
GANs are specific in the sense that they are constituted of two different components: a \textit{generator} and a \textit{discriminator}. Both are tightly coupled, as they compete to reach the learning target. The challenges for an efficient distribution are numerous; first, that coupling requires fine grained distribution strategies between workers, so that the bandwidth implied by the learning process remains acceptable. Second, the computational load on the workers has to be reasonable, as the purpose of distribution is also to gain efficiency regarding the training on a single GPU setup for instance. Lastly, as deep learning computation has shown not to be a deterministic process when considering the accuracy of the learned models facing various distribution scales \cite{Staleness-awareASGD_2015}, the accuracy of the model computed in parallel has to remain competitive.


\paragraph{Contributions} The contributions of this paper are: \\
\textit{(i)} to propose the first approach (MD-GAN) to distribute GANs over a set of worker machines. In order to provide an answer to the computational load challenge on workers, we remove half of their burden by having a single generator in the system, hosted by the parameter server. This is made possible by a peer-to-peer like communication pattern between the discriminators spread on the workers.\\
\textit{(ii)} to compare the learning performance of MD-GAN with regards to both the baseline learning method (\ie on a standalone server) and an adaption of \textit{federated learning} to GANs \cite{FederatedLearning_premices}. This permits head to head comparisons regarding the accuracy challenge.\\
\textit{(iii)} to experiment MD-GAN and the two other competitors on the MNIST and CIFAR10 datasets, using GPUs. In addition to analytic expectations of communication and computing complexities, this sheds light on the advantages of MD-GAN, but also on the salient properties of the MD-GAN and federated learning approaches for the distribution of GANs.\\

\paragraph{Paper organization} 
In Section \ref{GAN}, we give general background on GANs.
Section \ref{comp_setup} presents the computation setup we consider, and presents an adaptation of federated learning to GANs.
Section \ref{MD-GAN} details the MD-GAN algorithm.
We experiment MD-GAN and its competitors in Section \ref{experiments}.
In Section \ref{relatex}, we review the related work. 
We finally discuss futureworks and conclude in Section \ref{conclusion}.

\section{Background on Generative Adversarial Networks}
\label{GAN}

The particularity of GANs as initially presented in \cite{gan} is that their training phase is \textit{unsupervised}, \ie no description labels are required to learn from the data.
A classic GAN is composed of two elements: a \textit{generator} \G and a \textit{discriminator} \D. Both are deep neural networks. The generator takes as input a noise signal (\eg random vectors of size $k$ where each entry follows a normal distribution $\mathcal{N}(0,1)$) and generates data with the same format as training dataset data (\eg a picture of 128x128 pixels and 3 color channels). The discriminator receives as input either some data from two sources: from the generator or from the training dataset. The goal of the discriminator is to guess from which source the data is coming from.
At the beginning of the learning phase, the generator generates data from a probability distribution and the discriminator quickly learns how to differentiate that generated data from the training data.
After some iterations, the generator learns to generate data which are closer to the dataset distribution. If it eventually turns out that the discriminator is not able to differentiate both, 
this means that the generator has learned the distribution of the data in the training dataset (and thus has learned an unlabeled dataset in an unsupervised way).


Formally, let a given training dataset be included in the data space $X$, where $\bm{x}$ in that dataset follows a distribution probability $P_{\text{data}}$. A GAN, composed of generator \G and discriminator \D, tries to learn this distribution. 
As proposed in the original GAN paper \cite{gan}, we model the generator by the function $\mathcal{G}_{\bm{w}}: \R^\ell \longrightarrow  X $, where $\bm{w}$ contains the parameters of its DNN $\mathcal{G}_{\bm{w}}$ and $\ell$ is fixed. Similarly, we model  the discriminator by the function $\mathcal{D}_{\bm{\theta}}: X \longrightarrow  \left[0,1 \right]$ where $\mathcal{D}_{\bm{\theta}}(\bm{x})$ is the probability that $\bm{x}$ is a data from the training dataset, and $\bm{\theta}$ contains the parameters of the discriminator $\mathcal{D}_{\bm{\theta}}$. Writing $\log$ for the logarithm to the base $2$, the learning consists in finding the parameters $\bm{w}^*$ for the generator:
$$\bm{w}^* = \argmin_{\bm{w}}
    \max_{\bm{\theta}} 
        (A_{\bm{\theta}}
        + 
        B_{\bm{\theta},\bm{w}}), \mbox{ with}$$
$$A_{\bm{\theta}} = \mathbb{E}_{\bm{x} \sim P_{\text{data}}} 
            \left[
                \log \mathcal{D}_{\bm{\theta}}\left(\bm{x}\right)
            \right] \mbox{ and}$$
$$B_{\bm{\theta},\bm{w}} = \mathbb{E}_{\bm{z}\sim \bm{{\cal N}}_\ell}
            \left[
                \log 
                    \left(
                        1-\mathcal{D}_{\bm{\theta}} 
                        \left( 
                            \mathcal{G}_{\bm{w}}(\bm{z}
                        \right)
                    \right)
            \right],$$
\noindent where $\bm{z}\sim \bm{{\cal N}}_\ell$ means that each entry of the $\ell$-dimensional 
random vector $\bm{z}$ follows a normal distribution with fixed parameters. 
In this equation, \D adjusts its parameters $\bm{\theta}$ to maximize $A_{\bm{\theta}}$, \ie the expected good classification on real data and $B_{\bm{\theta},\bm{w}}$, the expected good classification on generated data. \G adjusts its parameters $\bm{w}$ to minimize $B_{\bm{\theta},\bm{w}}$ ($\bm{w}$ does not have impact on $A$), which means that it tries to minimize the expected good classification of \D on generated data. The learning is performed by iterating two steps, named the discriminator learning step and the generator learning step, as described in the following.

\subsubsection{Discriminator learning} \label{disclearning} The first step consists in learning $\bm{\theta}$ given a fixed $\mathcal{G}_{\bm{w}}$. The goal is to approximate the parameters $\bm{\theta}$ which maximize $A_{\bm{\theta}}+B_{\bm{\theta},\bm{w}}$ with the actual $\bm{w}$. This step is performed by a gradient descent (generally using the Adam optimizer \cite{adam}) of the following discriminator error function $J_{disc}$ on parameters $\bm{\theta}$: 
\begin{equation*}
J_{disc}(X_{r},X_{g}) = \widetilde{A}(X_{r}) + \widetilde{B}(X_{g}), \mbox{ with}
\end{equation*}
$$ \widetilde{A}(X_{r}) =  \frac{1}{b}\sum_{\bm{x}\in X_{r}} \hspace{-0.15cm} \log(\mathcal{D_{\bm{\theta}}}(\bm{x}));
\widetilde{B}(X_{g}) = \frac{1}{b}\sum_{\bm{x}\in X_{g}} \hspace{-0.15cm} \log(1-\mathcal{D_{\bm{\theta}}}(\bm{x})),$$
\noindent where $X_{r}$ is a batch of $b$ real data drawn randomly from the training dataset and $X_{g}$ a batch of $b$ generated data from \G. In the original paper \cite{gan}, the authors propose to perform few gradient descent iterations to find a good $\bm{\theta}$ against the 
fixed~$\mathcal{G}_{\bm{w}}$.

\subsubsection{Generator learning} The second step consists in adapting $\bm{w}$ to the new parameters $\bm{\theta}$. As done for step 1), it is performed by a gradient descent of the following error function $J_{gen}$ on generator parameters $\bm{w}$:
\begin{align*}
J_{gen}(Z_{g}) & = \widetilde{B}\left(\{\mathcal{G}_{\bm{w}}(\bm{z}) | \bm{z} \in Z_{g}\}\right) \\
& = \frac{1}{b} \sum_{\bm{x}\in \{\mathcal{G}_{\bm{w}}(\bm{z}) | \bm{z} \in Z_{g}\}}  \log(1-\mathcal{D}_{\bm{\theta}}(\bm{x})) \\
& = \frac{1}{b} \sum_{\bm{z}\in Z_{g}}  \log(1-\mathcal{D}_{\bm{\theta}}(\mathcal{G}_{\bm{w}}(\bm{z})))
\end{align*}

\noindent where $Z_g$ is a sample of $b$ $\ell-$dimensional random vectors generated from $\bm{{\cal N}}_\ell$. Contrary to discriminator learning step, this step is performed only once per iteration. 

By iterating those two steps a significant amount of times with different batches (see \eg \cite{gan} for convergence related questions), the GAN ends up with a $\bm{w}$ which approximates $\bm{w}^*$ well. Such as for standard deep learning, guarantees of convergence are weak \cite{principledMethodsGAN}. Despite this very recent breakthrough, there are lots of alternative proposals to learn a GAN (\eg more details can be found in \cite{wgan,acgan}, and\cite{ImprovedGAN}).


\section{Distributed computation setup for GANs}
\label{comp_setup}
Before we present MD-GAN in the next Section, we introduce the distributed computation setup considered in this paper, and an adaptation of federated learning to GANs.

\paragraph{Learning over a spread dataset} We consider the following setup. $N$ workers (possibly from several datacenters \cite{gaia}) are each equipped with a local dataset composed of $m$ samples (each of size $d$) from the same probability distribution $P_{\text{data}}$ (\eg requests to a voice assistant, holiday pictures). Those local datasets will remain in place (\ie will not be sent over the network). We denote by $\mathcal{B}=\bigcup_{n=1}^N \mathcal{B}_n$ the entire dataset, with $\mathcal{B}_n$ the dataset local to worker $n$. 
We assume in the remaining of the paper that the local datasets are $i.i.d.$ on workers, that is to say that there are no bias in the distribution of the data on one particular worker node.

\begin{figure*}[h!]
\centering
\includegraphics[width=7.4in]{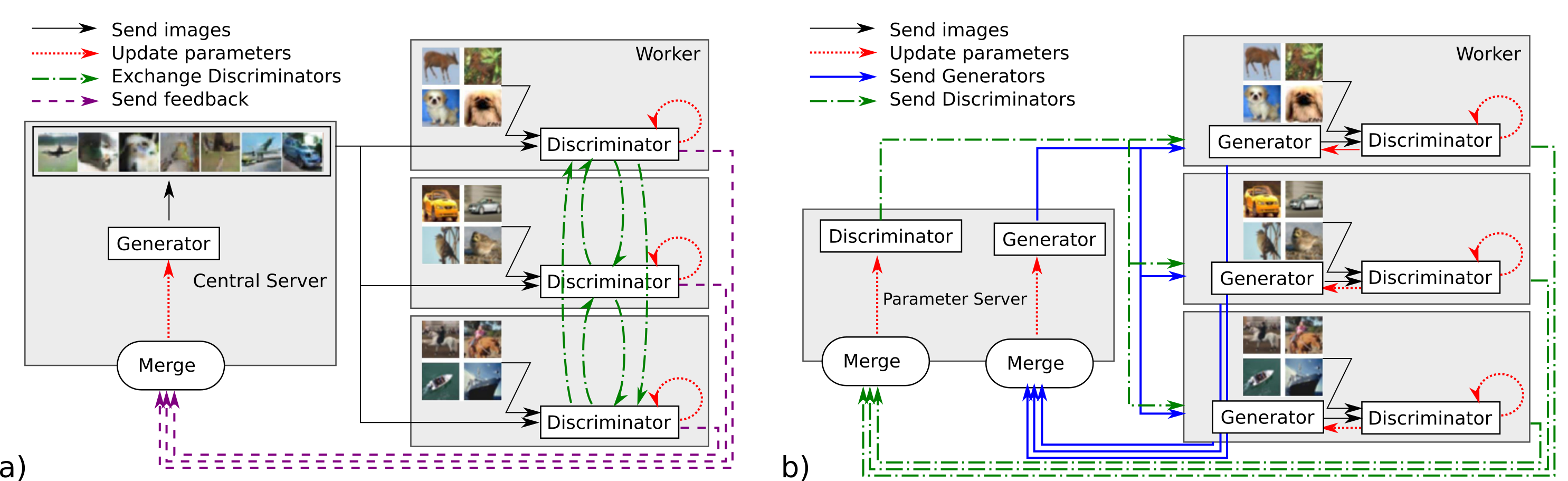}
\caption{The two proposed competitors for the distribution of GANs: a) The MD-GAN communication pattern, compared to b) FL-GAN (federated learning adapted to GANs). MD-GAN leverages a single generator, placed on the server; FL-GAN uses generators on the server and on each worker. MD-GAN swaps discriminators between workers in a peer-to-peer fashion, while in FL-GAN they stay fixed and are averaged by the server upon reception from the workers.}
\label{fig:arc}
\end{figure*}

The assumption on the fix location of data shares is complemented by the use of the parameter server framework we are now presenting.

\paragraph{The parameter server framework} Despite the general progress of distributed computing towards serverless operation even in datacenters (\eg use of the \textit{gossip} paradigm as in Dynamo \cite{dynamo} back in 2007), the case of deep learning systems is specific.
Indeed, the amounts of data required to train a deep learning model, and the very iterative nature of the learning tasks (learning on batches of data, followed by operations of back-propagations) makes it necessary to operate in a parallel setup, with the use of a central server.
Introduced by Google in 2012~\cite{DistBelief_DownpourSGD_2012}, the \textit{parameter server} framework uses \textit{workers} for parallel processing, while one or a few central servers are managing shared states modified by those workers (for simplicity, in the remaining of the paper, we will assume the presence of a single central server). 
The method aims at training the same model on all workers using their given data share, and to synchronize their learning results with the server at each iteration, so that this server can update the model parameters.

Note that more distributed approaches for deep learning, such as gossip-based computation \cite{gossipDL,Hardy:2018:GGP:3286490.3286563}, have not yet proven to work efficiently on the data scale required for modern applications; we thus leverage a variant of parameter server framework as our computation setup.

\paragraph{FL-GAN: adaptation of federated learning to GANs}
\label{fl}
By the design of GANs, a generator and a discriminator are two separate elements that are yet tightly coupled; this fact makes it nevertheless possible to consider adapting a known computation method, that is generally used for training a single deep neural network.\footnote{We note that more advanced GAN techniques such as those by Wang et al. \cite{DBLP:journals/corr/WangZW16} or by Tolstikhin et al. \cite{NIPS2017_7126} might also be distributed and serve as baselines; yet this distribution requires a full redesign of the proposed protocols, and is thus out of the scope of this paper.}
Federated learning \cite{FederatedLearning_Comm_2016} proposes to train a machine learning model, and in particular a deep neural network, on a set of workers. It follows the parameter server framework, with the particularity that workers perform numerous local iterations between each communication to the server (\ie a round), instead of sending small updates. All workers are not necessarily active at each round; to reduce conflicting updates, all active workers synchronize their model with the server at the beginning of each round.

In order to compare MD-GAN to a federated learning type of setup, we propose an adapted version of federated learning to GANs. This adaptation considers the discriminator \D and generator \G on each worker as one computational object to be treated atomically. 
Workers perform iterations locally on their data and every $E$ \textit{epochs} (\ie each worker passes $E$ times the data in their GAN) they send the resulting parameters to the server. The server in turn averages the \G and \D parameters of all workers, in order to send updates to those workers at the next iteration. We name this adapted version FL-GAN; it is depicted by Figure \ref{fig:arc} b).









We now detail MD-GAN, our proposal for the learning of GANs over workers and their local datasets.

\section{The MD-GAN algorithm}
\label{MD-GAN}

\subsection{Design rationale} To diminish computation on the workers, we propose to operate with a single \G, hosted on the  server\footnote{In that regard, MD-GAN do not fully comply with the parameter server model, as the workers do not compute and synchronize to the same model architecture hosted at the server. Yet, it leverages the parallel computation and the iterative nature of the learning task proposed by the the parameter server framework.}. 
That server holds parameters $\bm{w}$ for \G; data shares are split over workers. 
To remove part of the burden from the server, discriminators are solely hosted by workers, and  
move in a peer-to-peer fashion between them. Each worker $n$ starts with its own discriminator $\mathcal{D}_n$ with parameters $\bm{\theta}_n$. Note that the architecture and initial parameters of $\mathcal{D}_n$ could be different on every worker $n$; for simplicity, we assume that they are the same. This architecture is presented on 
 Figure \ref{fig:arc} a).
 
The goal for GANs is to train generator \G using $\mathcal{B}$. In MD-GAN, the \G on the server is trained using the workers and their local shares. It is a 1-versus-N game where \G faces all $\mathcal{D}_n$, \ie \G tries to generate data considered as real by all workers. Workers use their local datasets $\mathcal{B}_n$ to differentiate generated data from real data. 
Training a generator is an iterative process; in MD-GAN, a \textit{global learning iteration} is composed of four steps:
\begin{itemize}
\item The server generates a set $K$ of $k$ batches $K=\{X^{(1)},\dots,X^{(k)}\}$, with $k\leq N$. Each $X^{(i)}$ is composed of $b$ data generated by \G. 
The server then selects, for each worker $n$, two distinct batches, say $X^{(i)}$ and $X^{(j)}$, which are sent to worker $n$ and locally renamed as $X_n^{(g)}$ and $X_n^{(d)}$.
The way in which the two distinct batches are selected is discussed in Section~\ref{batchesSelection}.
\item Each worker $n$ performs $L$ learning iterations on its discriminator $\mathcal{D}_n$ (see Section \ref{disclearning}) using $X_n^{(d)}$ and $X_n^{(r)}$, where $X_n^{(r)}$ is a batch of real data extracted locally from~$\mathcal{B}_n$.
\item Each worker $n$ computes an error feedback $F_n$ on $X_n^{(g)}$ by using $\mathcal{D}_n$ and sends this error to the server. We detail in Section \ref{errorFeedBack} the computation of $F_n$. 
\item The server computes the gradient of $J_{gen}$ for its parameters $\bm{w}$ using all the $F_n$ feedbacks. It then updates its parameters with the chosen optimizer algorithm (\eg Adam \cite{adam}).
\end{itemize}
Moreover, every $E$ epochs, workers start a peer-to-peer swapping process for their discriminators, using function \textsc{Swap}().
The pseudo-code of MD-GAN, including those steps, is presented in Algorithm \ref{algo:mdgan}.

Note that extra workers can enter the learning task if they enter with a pre-trained discriminator (\eg a copy of another worker discriminator); we discuss worker failures in Section \ref{experiments}. 

\begin{table}[t]
\centering
\begin{tabular}{|c|p{2in}|}
\hline
Notation & \\
\hline
\G & Generator \\
\D & Discriminator \\
$N$ & Number of workers \\
$C$ & Central server \\
$W_n$ & Worker $n$ \\
$P_{\text{data}}$ & Data distribution \\
$P_{\mathcal{G}}$ & Distribution of generator \G \\
$\bm{w}$ (resp. $\bm{\theta}$) & Parameters of \G (resp. \D) \\
$w_i$ (resp. $\theta_i$) & $i$-th parameter of \G (resp. \D) \\
$\mathcal{B}$ & Distributed training dataset \\
$\mathcal{B}_n$ & Local training dataset on worker $n$ \\
$m$ & Number of objects in a local dataset $\mathcal{B}_n$\\
$d$ & Object size (\eg image in Mb) \\
$b$ & Batch size \\
$I$ & Number of training iterations \\
$K$ & The set of all batches $X^{(1)},\dots,X^{(k)}$ generated by \G during one iteration\\
$F_n$ & The error feedback computed by worker $n$\\
$E$ & Number of local epochs before swapping discriminators\\
\hline
\end{tabular}
\caption{Table of notations}
\label{tab:notations}
\end{table}

\subsection{The generator learning procedure (server-side)}

The server hosts generator \G with its associated parameters $\bm{w}$. Without loss of generality, this paper exposes the training of GANs for image generation; the server generates new images to train all discriminators and updates $\bm{w}$ using error feedbacks. 

\subsubsection{Distribution of generated batches}
\label{batchesSelection}
Every global iteration, \G generates a set of $k$ batches $K=\{X^{(1)}, \dots, X^{(k)}\}$ (with $k\leq N$) of size $b$. Each participating worker $n$ is sent two batches among $K$, $X^{(g)}_n$ and $X^{(d)}_n$. 
This two-batch generation design is required, for the computation of the gradients for both \D and \G on separate data (such as for the original GAN design \cite{gan}).
A possible way to distribute the $X^{(i)}$ among the $N$ workers could be to set 
$X^{(g)}_n = X^{((n\mod k)+1)}$ and $X^{(d)}_n = X^{(((n+1)\mod k)+1)}$ for 
$n=1,\ldots,N$. 

\subsubsection{Update of generator parameters}
\label{errorFeedBack}
Every global iteration, the server receives the error feedback $F_n$ from every worker $n$, corresponding to the error made by \G on $X^{(g)}_n$. More formally, $F_n$ is composed of $b$ vectors $\{ \bm{e}_{n_1},\dots,\bm{e}_{n_b} \}$, where $\bm{e}_{n_i}$ is given by

\begin{equation*}
\bm{e}_{n_i} = \frac{\partial \widetilde{B}(X^{(g)}_n)}{\partial \bm{x}_{i}},
\end{equation*}

\noindent with $\bm{x}_i$ the $i$-th data of batch $X^{(g)}_n$. The gradient 
$\Delta \bm{w} = \partial \widetilde{B}\left(\cup_{n=1}^N X^{(g)}_n\right)/\partial \bm{w}$ is deduced from all $F_n$ as 

\begin{equation*}
\Delta w_j = \frac{1}{Nb} \sum_{n=1}^N \sum_{\bm{x}_i\in X_n^{(g)} } \bm{e}_{n_i}
\frac{\partial \bm{x}_i}{\partial w_j},
\end{equation*}

\noindent with $\Delta w_j$ the $j$-th element of $\Delta \bm{w}$. The term 
$\partial \bm{x}_i/\partial w_j$ is computed on the server. Note that $\cup_{n=1}^N X_n^{(g)} = \{ \mathcal{G}_w(\bm{z}) | \bm{z} \in Z_g \}$. Minimizing $\widetilde{B}\left(\cup_{n=1}^N X^{(g)}_n\right)$ is thus equivalent to minimize $J_{gen}(Z_g)$. Once the gradients are computed, the server is able to update its parameters $\bm{w}$.
We thus choose to merge the feedback updates through an averaging operation, as it is the most common way to aggregate updates processed in parallel \cite{Hogwild_NIPS2011,DistBelief_DownpourSGD_2012,AsyncSGD_NIPS2015,SSGD}.
Using the Adam optimizer \cite{adam}, parameter $w_i \in \bm{w}$ at iteration $t$, denoted by $w_i(t)$ here, is computed as follows:



\begin{equation*}
w_j(t) = w_j(t-1) + \text{Adam(} \Delta w_j \text{)} ,
\end{equation*}
where the Adam optimizer is the function which computes the update given the gradient $\Delta w_j$.


\begin{algorithm}[t!]
\begin{algorithmic}[1]

\Procedure{Worker}{$C,\mathcal{B}_n,I,L,b$}
    \State Initialize $\bm{\theta}_n$ for $\mathcal{D}_n$
	\For{$i \gets 1$ \textbf{to} $I$}
	    \State $X_n^{(r)} \gets $ \Call{Samples}{$\mathcal{B}_n,b$}
	    \State $X^{(g)}_n, X^{(d)}_n\gets $ \Call{ReceiveBatches}{$C$}
	    \For{$l \gets 0$ \textbf{to} $L$}
	        \State $\mathcal{D}_n \gets $\Call{DiscLearningStep}{$J_{disc}$,$\mathcal{D}_n$} 
	    \EndFor
	    \State $F_n \gets \{\frac{\partial \widetilde{B}(X^{(g)}_n)}{\partial \bm{x}_{i}} | \bm{x}_i \in X_n^{(g)} \}$
	    \State \Call{Send}{$C,F_n$} \Comment{Send $F_n$ to server}
	    \If{$i\mod (\frac{mE}{b})=0$}
	        \State $\mathcal{D}_n \gets $\Call{Swap}{$\mathcal{D}_n$}
	    \EndIf
	\EndFor
\EndProcedure\\

\Procedure{Swap}{$\mathcal{D}_n$}
\State $W_l\gets $ \Call{GetRandomWorker}{}
\State \Call{Send}{$W_l,\mathcal{D}_n$}\Comment{Send $\mathcal{D}_n$ to worker $W_l$.}
\State $\mathcal{D}_n \gets$ \Call{ReceiveD}{} \Comment{Receive a new discriminator from another worker.}
\State Return $\mathcal{D}_n$
\EndProcedure\\

\Procedure{Server}{$k$,$I$} \Comment{Server C}
\State Initialize $\bm{w}$ for \G
	\For{$i \gets 1$ \textbf{to} $I$}
	    \For{$j \gets 0$ \textbf{to} $k$}
	        \State $Z_j \gets $\Call{GaussianNoise}{$b$}
	        \State $X^{(j)} \gets \{ \mathcal{G}_{\bm{w}}(\bm{z}) | \bm{z} \in Z_j \}$
	    \EndFor
        \State $X_1^{(d)},\dots,X_n^{(d)} \gets$ \Call{Split}{$X^{(1)},\dots,X^{(k)}$}
        \State $X_1^{(g)},\dots,X_n^{(g)} \gets$ \Call{Split}{$X^{(1)},\dots,X^{(k)}$}
        \For{$n \gets 1$ \textbf{to} $N$}
            \State \Call{Send}{$W_n,(X_{n}^{(d)},X_n^{(g)})$}
        \EndFor
        
        \State ${F_1,\dots,F_N} \gets $ \Call{GetFeedbackFromWorkers}{}
        \State Compute $\Delta \bm{w}$ according to ${F_1,\dots,F_N}$
        
        \For{$w_i\in \bm{w}$}
            \State $w_i \gets w_i +$\Call{Adam}{$\Delta w_i$} 
        \EndFor
    \EndFor
\EndProcedure

\end{algorithmic}
\caption{MD-GAN algorithm}
\label{algo:mdgan}
\end{algorithm}

\subsubsection{Workload at the server}

Placing the generator on the server increases its workload. It generates $k$ batches of $b$ data using \G during the first step of a global iteration, and then receives $N$ error feedbacks of size $bd$ in the third step. 
The batch generation requires $kbG_{op}$ floating point operations (where $G_{op}$ is the number of floating operations to generate one data object with \G) and a memory of $kbG_a $ (with $G_a$ the number of neurons in \G). For simplicity, we assume that $G_{op}= O(|\bm{w}|)$ and that $G_a$ = $O(|\bm{w}|)$.
Consequently the batch generation complexity is 
$O(kb|\bm{w}|)$.
The merge operation of all feedbacks $F_n$ and the gradient computations imply a memory and computational complexity of $O(b(dN + k|\bm{w}|))$.

\subsubsection{The complexity vs. data diversity trade-off}
\label{tradeoff}

At each global iteration, the server generates $k$ batches, with $k<N$. If $k=1$, all workers receive and compute their feedback on the same training batch. This reduces the diversity of feedbacks received by the generator but also reduces the server workload. If $k=N$, each worker receives a different batch, thus no feedback has conflict on some concurrently processsed data. In consequence, there is a trade-off regarding the generator workload: because $k=N$ seems cumbersome, we choose $k=1$ or $k= \lfloor \log(N) \rfloor$ for the experiments, and assess the impact of those values on final model performances.

\subsection{The learning procedure of discriminators (worker-side)}

Each worker $n$ hosts a discriminator $\mathcal{D}_n$ and a training dataset $\mathcal{B}_n$. It receives batches of generated images split in two parts: $X_n^{(d)}$ and $X_n^{(g)}$. The generated images $X_n^{(d)}$ are used for training $\mathcal{D}_n$ to discriminate those generated images from real images. 
The learning is performed as a classical deep learning operation on a standlone server \cite{gan}. A worker $n$ computes the gradient $\Delta \bm{\theta}_n$ of the error function $J_{disc}$ applied to the batch of generated images $X^{(d)}_n$, and a batch or real image $X^{(r)}_n$ taken from $\mathcal{B}_n$. As indicated in Section \ref{disclearning}, this operation is iterated $L$ times. 
The second batch $X_n^{(g)}$ of generated images is used to compute the error term $F_n$ of generator \G. Once computed, $F_n$ is sent to the server for the computation of gradients $\Delta \bm{w}$.

\subsubsection{The swapping of discriminators}
Each discriminator $n$ solely uses $\mathcal{B}_n$ to train its parameters $\bm{\theta}_n$. 
If too many iterations are performed on the same local dataset, the discriminator tends to over specialize (which  decreases its capacity of generalization). This effect, called \textit{overfitting}, is avoided in MD-GAN by swapping the parameters of discriminators $\bm{\theta}_n$ between workers after $E$ epochs.
The swap is implemented in a gossip fashion, by choosing randomly for every worker another worker to send its parameters to.

\subsubsection{Workload at workers}

The goal of MD-GAN is to reduce the workload of workers without moving data shares out of their initial location. Compared to our proposed adapted federated learning method FL-GAN, the generator task is deported on the server. Workers only have to handle their discriminator parameters $\theta_n$ and to compute error feedbacks after $L$ local iterations. Every global iteration, a worker performs $2bD_{op}$ floating point operations (where $D_{op}$ is the number of floating point operations for a feed-forward step of \D for one data object). The memory used at a worker is $O(|\bm{\theta}|)$. 

\begin{table}[t!]
\centering
\begin{tabular}{|c|c|c|}
\hline
 & FL-GAN & MD-GAN\\
\hline
Computation C & $O(I b N(|\bm{w}|+|\bm{\theta}|)/(mE))$ & $O(Ib(dN+ k |\bm{w}|))$ \\ 
Memory C & $O(N(|\bm{w}|+|\bm{\theta}|))$ & $O(b(dN + k |\bm{w}|)$ \\
\rowcolor{gray} Computation W & $O(I b (|\bm{w}|+ |\bm{\theta}|))$ & $O(I b |\bm{\theta}|)$ \\ 
\rowcolor{gray} Memory W & $O(|\bm{w}|+|\bm{\theta}|)$ & $O(|\bm{\theta}|)$ \\
\hline
\end{tabular}
\caption{Computation complexity and memory for MD-GAN and adapted federated learning to GANs. The rows in grey highlight the reduction by a factor of two for MD-GAN on workers.}
\label{tab:complexities}
\end{table}

\subsection{The characteristic complexities of MD-GAN}

\subsubsection{Communication complexity}
In the MD-GAN algorithm there are three types of communications:
\begin{itemize}
\item Server to worker communication: the server sends its $k$ batches of generated images to workers at the beginning of global iterations. The number of generated images is $k b$ (with $k \leq N$), but only two batches are sent per worker. The total communication from the server is thus $2bd N$ (\ie $2bd$ per worker). 
\item Worker to server communications: after computing the generator errors on $X^{(g)}_n$, all workers send their error term $F_n$ to the server. The size of error term is $b d$ per worker, because solely one float is required for each feature of the data. 
\item Worker to worker communications: after $E$ local epochs, each discriminator parameters are swapped. Each worker sends a message of size $| \bm{\theta}_n |$, and receive a message of the same size (as we assume for simplicity that discriminator models on workers have the same architecture).

\end{itemize}

Communication complexities are summarized in Table \ref{tab:com}, for both MD-GAN and FL-GAN.
Table \ref{tab:com_exp} instantiates those complexities with the actual quantities of data measured for the experiment on the CIFAR10 dataset.
The first observation is that MD-GAN requires server to workers communication at every iteration, while FL-GAN performs $mE/b$ iterations in between two communications. Note that the size of workers-server communications depends on the GAN parameters ($\bm{\theta}$ and $\bm{w}$) for FL-GAN, whereas it depends on the size of data objects and on the batch size in MD-GAN. It is particularly interesting to choose a small batch sizes, especially since it is shown by Gupta et al. \cite{Rudra_2015} that in order to hope for good performances in the parallel learning of a model (as discriminators in MD-GAN), the batch size should be inversely proportional to the number of workers $N$. When the size of data is around the number of parameters of the GAN (such as in image applications), the MD-GAN communications may be expensive. For example, GoogLeNet \cite{GoogleNet_2015} analyzes images of $224 \times 224$ pixels in RGB ($150,528$ values per data) with less than $6.8$ millions of parameters. 

\begin{table}
\begin{center}
\begin{tabular}{|c|c|c|}
\hline
Communication type & FL-GAN & MD-GAN \\
\hline
 C$\rightarrow$W (C) & $N(\bm{\theta}+\bm{w})$ & $bdN$ \\
 C$\rightarrow$W (W) & $\bm{\theta}+\bm{w}$ & $b d$ \\
 W$\rightarrow$C (W) & $\bm{\theta}+\bm{w}$ & $b d$ \\
 W$\rightarrow$C (C) & $N(\bm{\theta}+\bm{w})$ & $bdN$ \\
Total \# C$\leftrightarrow$D & $I b/(m E)$ & $I$ \\
 W$\rightarrow$W (W) & - & $\bm{\theta}$ \\
Total \# W$\leftrightarrow$W & - & $Ib/(mE)$ \\
\hline
\end{tabular}
\caption{Communication complexities for both MD-GAN and FL-GAN. C and W stand for the central server and the workers, respectively.} 
 \label{tab:com}
\end{center}
\end{table}

\begin{table}
\begin{center}
\begin{tabular}{|c|c|c|c|c|}
\hline 
Communication type & FL-GAN & FL-GAN & MD-GAN & MD-GAN \\
 & $b=10$ & $b=100$ & $b=10$ & $b=100$ \\
\hline
 C$\rightarrow$W (C) & 175 MB & 175 MB & 2.30 MB & 23.0 MB \\
 C$\rightarrow$W (W) & 17.5 MB & 17.5 MB & 0.23 MB & 2.30 MB \\
 W$\rightarrow$C (W) & 17.5 MB & 17.5 MB & 0.23 MB & 2.30 MB \\
 W$\rightarrow$C (C) & 175 MB & 175 MB & 2.30 MB & 23.0 MB \\
Total \# C$\leftrightarrow$W & 100 & 1,000 & 50,000 & 50,000\\
 W$\rightarrow$W (W) & - & - & 6.34 MB & 6.34 MB MB \\
Total \# W$\leftrightarrow$W & - & - & 100 & 1,000\\
\hline

\end{tabular}
\caption{Example of communication costs for both MD-GAN and FL-GAN, in the CIFAR10 experiment with 10 workers.}
\label{tab:com_exp}
\vspace{-0.3cm}
\end{center}
\end{table}

\begin{figure}
\centering
\includegraphics[width=3.3in]{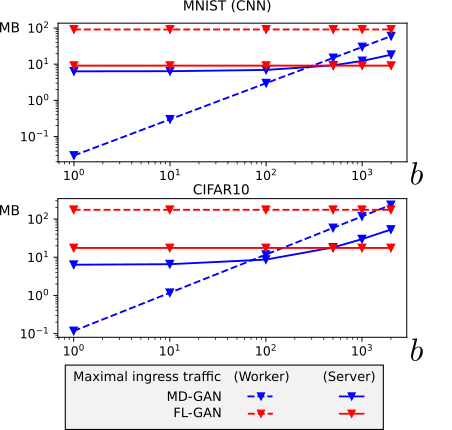}
\caption{Maximal ingress traffic, per communication, for two types of GANs (for MD-GAN and FL-GAN).}
\label{fig:ingress}
\end{figure}

We plotted on Figure \ref{fig:ingress} an analysis of the maximum ingress traffic ($x$-axis) of the FL-GAN and MD-GAN schemes, for a single iteration, and depending on chosen batch size ($y$-axis).
This corresponds for FL-GAN to a worker-server communication, and for MD-GAN for both worker-server and worker-worker communications during an iteration. 
Plain lines depict the ingress traffic at workers, while dotted lines the traffic at the server; these quantities can help to dimension the network capabilities required for the learning process to take place. Note the log-scale on both axis.

As expected the FL-GAN traffic is constant, because the communications depends only on the model sizes that constitute the GAN; it indicates a target upper bound for the efficiency of MD-GAN.
MD-GAN lines crossing FL-GAN is indicating more incurring traffic with increasing batch sizes. A global observation is that MD-GAN is competitive for smaller batch sizes, yet in the order of hundreds of images (here of less than around $b=550$ for MNIST and $b=400$ for CIFAR10).

\subsubsection{Computation complexity}

The goal of MD-GAN is to remove the generator tasks from workers by having a single one at the server. During the traning of MD-GAN, the traffic between workers and the server is reasonable (Table \ref{tab:com}). The complexity gain on workers  in term of memory and computation depends on the architecture of \D; it is generally half of the total complexity because \G and \D are often similar. The consequence of this single generator-based algorithm is more frequent interactions between workers and the server, and the creation of a worker-to-worker traffic.  
The overall operation complexities are summarized and compared in Table \ref{tab:complexities}, for both MD-GAN and FL-GAN;
the Table indicates a workload of half the one of FL-GAN on workers.

\section{Experimental evaluation}
\label{experiments}

We now analyze empirically the convergence of MD-GAN and of competing approaches. 

\subsection{Experimental setup}


Our experiments are using the Keras framework with the Tensorflow backend. We emulated workers and the server on GPU-based servers equipped of two Intel Xeon Gold 6132 processor, 260 GB of RAM and four NVIDIA Tesla M60 GPUs or four NVIDIA Tesla P100 GPUs. This setup allows for a training of GANs that is identical to a real distributed deployement, as computation order of interactions for Algorithm \ref{MD-GAN} are preserved. This choice for emulation is thus oriented towards a tighter control for the environment of competing approaches, to report more precise head to head result comparisons; raw timing performances of learning tasks are in this context inaccessible and are left to futurework.

\paragraph{Datasets}
We experiment competing approaches on two classic datasets for deep learning: MNIST \cite{lecun1998mnist} and CIFAR10 \cite{cifar10}. MNIST is composed of a training dataset of $60,000$ grayscale images of $28 \times 28$ pixels representing handwritten digits and another test dataset of $10,000$ images. Theses two datasets are composed respectively of $6,000$ and $1,000$ images for each digits. 
CIFAR10 is composed of a training set $50,000$ RGB images of $32 \times 32$ pixels representing the followings $10$ classes: airplane, automobile, bird, cat, deer, dog, frog, horse, ship, truck. CIFAR10 has a test dataset of $10,000$ images.

\paragraph{GAN architectures}

In the experiments, we train a classical type of GAN named ACGAN \cite{acgan}. We experiment with three different architectures for \G and \D: a multi-layer based architecture (MLP), a convolutional neural network based architecture (CNN) for MNIST and a CNN-based architecture for CIFAR10. Their characteristics are:
\begin{itemize}
\item In the MLP-based architecture for MNIST, \G and \D are composed of three fully-connected layers each. \G layers contain respectively $512$, $512$ and $784$ neurons, and \D layers contain $512$, $512$ and $11$ neurons. The total number of parameters is $716,560$ for \G and $670,219$ for \D.
\item In the CNN-based architecture for MNIST, \G is composed of one full-connected layer of $6,272$ neurons and two transposed convolutional layers of respectively $32$ and $1$ kernels of size $5\times5$. \D is composed of six convolutional layers of respectively $16$, $32$, $64$, $128$, $256$ and $512$ kernels of size $3\times3$, a mini-batch discriminator layer \cite{ImprovedGAN} and one full-connected layer of $11$ neurons. The total number of parameters is $628,058$ for \G and $286,048$ for \D.
\item In the CNN-based architecture for CIFAR10, \G is composed of one full-connected layer of $6,144$ neurons and three transposed convolutional layers of respectively $192$, $96$, and $3$ kernels of size $5\times5$. \D is composed of six convolutional layers of respectively $16$, $32$, $64$, $128$, $256$ and $512$ kernels of size $3\times3$, a mini-batch discriminator layer and one full-connected layer of $11$ neurons. The total number of parameters is $628,110$ for \G and $100,203$ for \D.
\end{itemize}

\begin{figure*}[t!]
\centering
\includegraphics[width=6.7in]{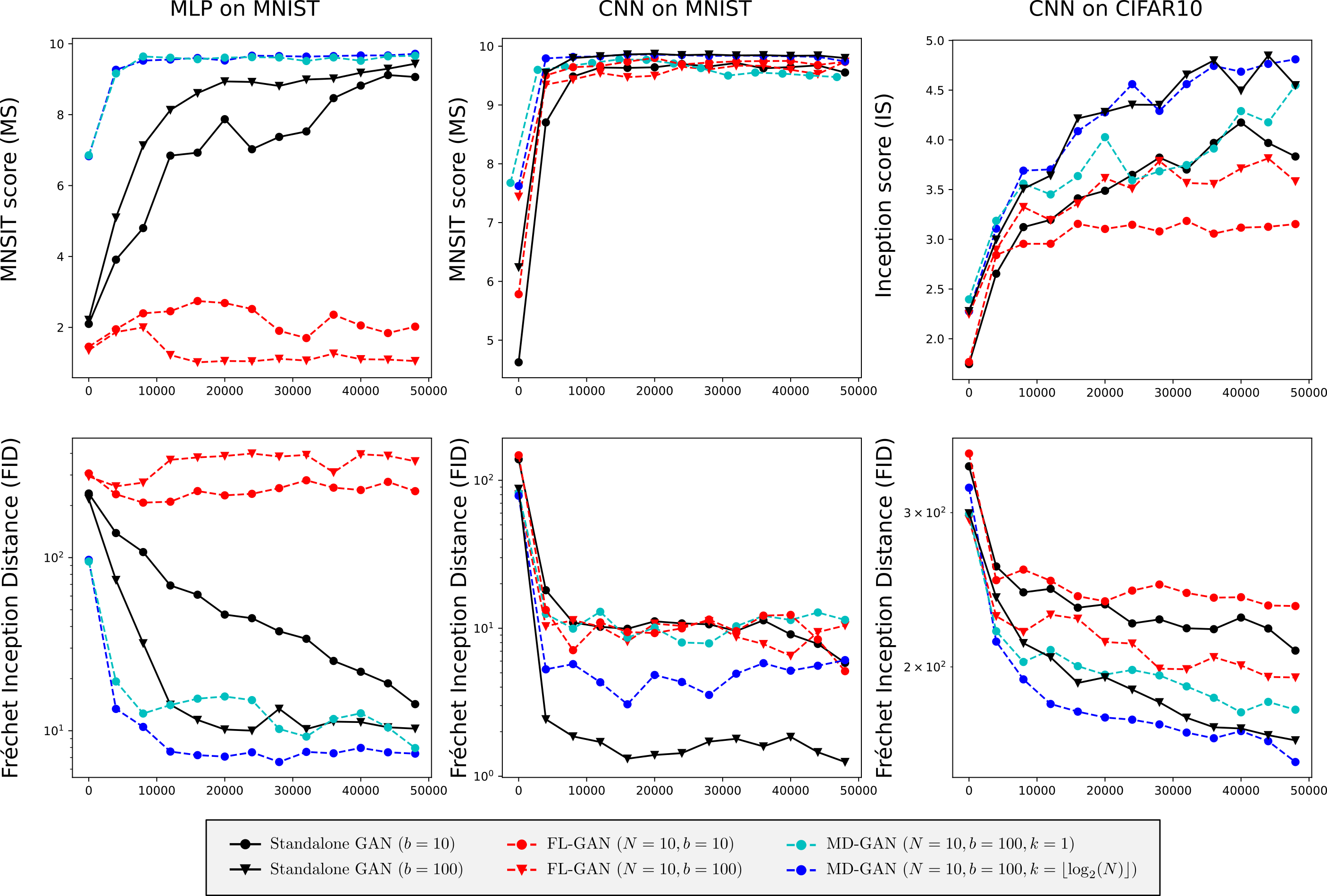}
\caption{MNIST score / Inception score (higher is better) and Fréchet Inception Distance (lower is better) for the three competing approaches, with regards to the number of iterations ($x$-axis).}
\label{fig:re1}
\end{figure*}

\paragraph{Metrics}
Evaluating generative models such as GANs is a difficult task. Ideally, it requires human judgment to assess the quality of the generated data. Fortunately, in the domain of GANs, interesting methods are proposed to simulate this human judgment. The main one is named the \textit{Inception Score} (we denote it by IS), and has been proposed by Salimans \textit{et al.} \cite{ImprovedGAN}, and shown to be correlated to human judgment. The IS consists to apply a pre-trained Inception classifier over the generated data. 
The Inception Score evaluates the confidence on the generated data classification (\ie generated data are well recognized by the Inception network), and on the diversity of the output (\ie generated data are not all the same). To evaluate the competitors on MNIST, we use the MNIST score (we name it MS), similar to the Inception score, but using a classifier adapted to the MNIST data instead of the Inception network. 
Heusel \textit{et al.} propose a second metric named the Fréchet Inception Distance (FID) in \cite{FID}. The FID measures a distance between the distribution of generated data $P_\mathcal{G}$ and real data $P_{\text{data}}$. It applies the Inception network on a sample of generated data and another sample of real data and supposes that their outputs are Gaussian distributions. The FID computes the Fréchet Distance 
between the Gaussian distribution obtained using generated data and the Gaussian distribution obtained using real data. As for the Inception distance, we use a classifier more adapted to compute the FID on the MNIST dataset. We use the implementation of the MS and FID available in Tensorflow\footnote{Code available at  \href{https://github.com/tensorflow/models/blob/master/research/gan/mnist/util.py}{https://github.com/tensorflow/models/blob/master/research/ gan/mnist/util.py}.}. 

\paragraph{Configurations of MD-GAN and competing approaches}

To compare MD-GAN to classical GANs, we train the same GAN architecture on a standalone server (it thus has access to the whole dataset $\mathcal{B}$). We name this baseline \textit{standalone-GAN} and parametrize it with two batch sizes $b=10$ and $b=100$. 

We run FL-GAN with parameters $E=1$ and $b=10$ or $b=100$; this parameter setting comes from the fact that $E=1$ and $b=10$ is one of the best configuration regarding computation complexity on MNIST, and because $b=50$ is the best one for performance per iteration \cite{FederatedLearning_premices} (having $b=100$ thus allows for a fair comparison for both FL-GAN and MD-GAN).
MD-GAN is run with also $E=1$; \ie for FL-GAN and MD-GAN, respective actions are taken after the whole dataset has been processed once.

For MD-GAN and FL-GAN, the training dataset is split equally over workers (images are sampled $i.i.d$).
We run two configurations of MD-GAN, one with $k=1$ and another with $k=\lfloor \log(N) \rfloor$, in order to evaluate the impact of the data diversify sent to workers. 
Finally, in FL-GAN, GANs over workers perform learning iterations (such as in the standalone case) during $1$ epoch, \ie until $\mathcal{D}_n$ processes all local data $\mathcal{B}_n$.  

We experimented with a number of workers $N \in [1,10,25,50]$; 
geo-distributed approaches such as Gaia \cite{gaia} or \cite{geo} also operate at this scale (where $8$ nodes \cite{geo} and $22$ nodes \cite{gaia} at maximum are leveraged).
All experiments are performed with $I=50,000$, \ie the generator (or the $N$ generators in FL-GAN) are updated $50,000$ times during a generator learning step. We compute the FID, MS and IS scores every $1,000$ iterations using a sample of $500$ generated data. The FID is computed using a batch of the same size from the test dataset. In FL-GAN, the scores are computed using the generator on the central server.

\subsection{Experiment results}

 We report the scores of all competitors, with regards to the iterations, on the Figure \ref{fig:re1}. The resulting curves are smoothed for readability.

\begin{figure}
\centering
\includegraphics[width=3.1in]{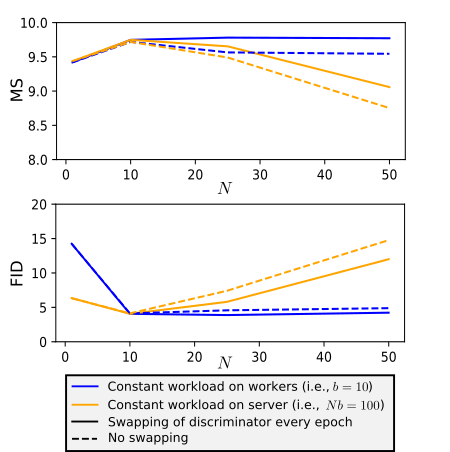}
\caption{MNIST score and Fréchet Inception Distance with regards to the varying number of workers for MD-GAN using the MLP model. Experiments include the disabling of the swapping processing for comparison purposes.}
\label{fig:scale}
\end{figure}

\begin{figure*}[t!]
\centering
\includegraphics[width=6.7in]{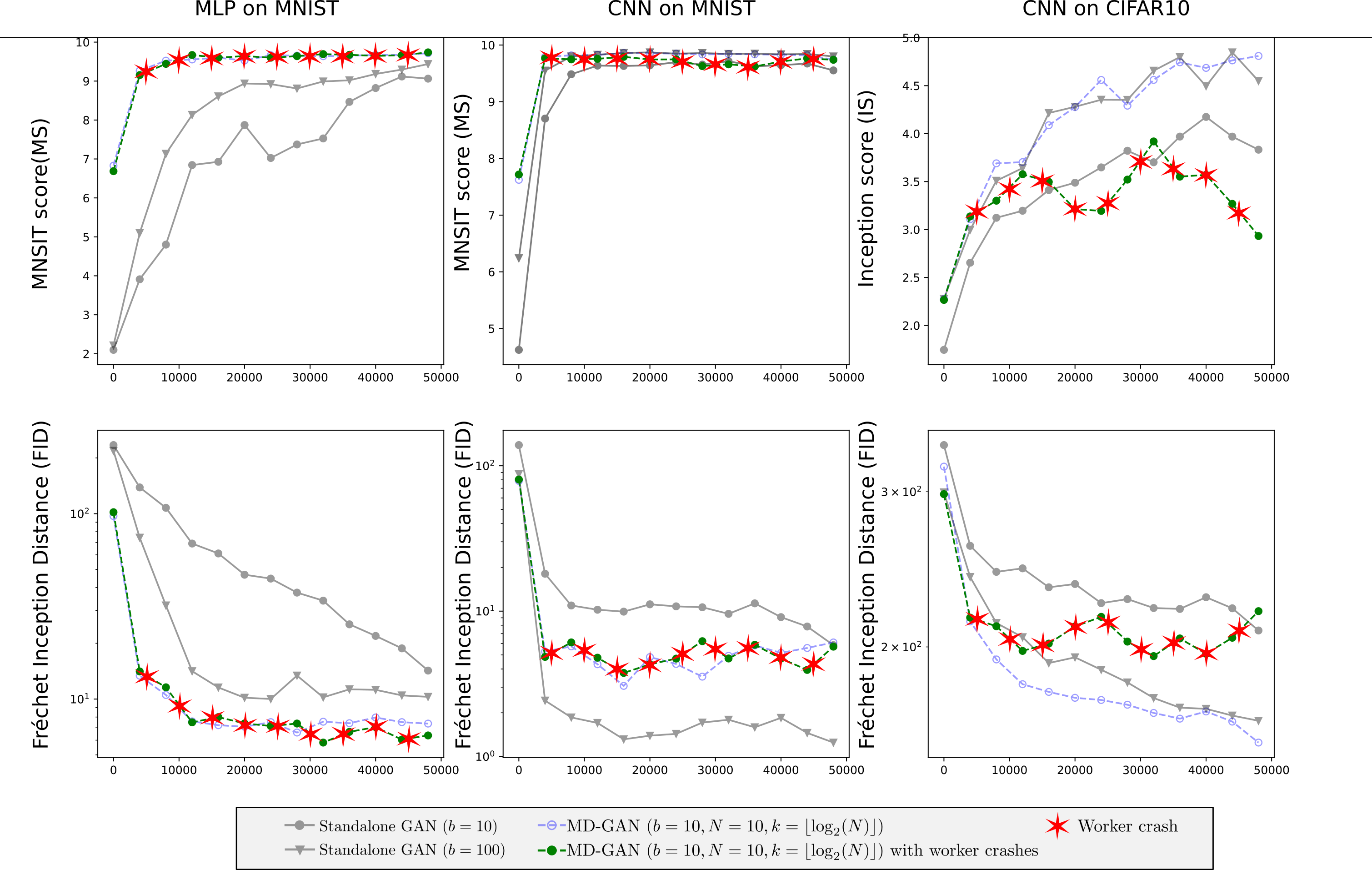}
\caption{MNIST score or Inception score and Fréchet Inception Distance over the number of iterations for MD-GAN with crash faults, compared to MD-GAN without any crash and to a standalone GAN.}
\label{fig:crashes}
\end{figure*}

\subsubsection{Competitor scores}
\label{comparison-scores}

The standalone GAN obtains better results with $b=100$, rather than with $b=10$. It is because the GAN sees more samples (real and generated data) per iteration when $b$ increases. 
When $b=10$ for MD-GAN, the total number of real data seen in all $\mathcal{B}_n$ is $100$ with $N=10$. This explains why MD-GAN obtains very similar scores than standalone GAN with $b=100$ (except with the CNN on MNIST). 
We note that, as highlighted in discussion in Section \ref{tradeoff}, the hyper-parameter $k$ has a significant impact on the learning process. The more the data diversity sent by the server to workers, the higher the generator scores.

For the experiments on MLP, FL-GAN does not converge, whereas MD-GAN has better scores (FID and MS) than the standalone competitor. We propose a multi-discriminator vs one generator game; some recent works \cite{MGAN} have shown that some central strategies based on one generator and multiple discriminators, or a \textit{mixture} of generators and one discriminator \cite{mGGAN}, can as well exceed the performances of a standalone GAN.
In the CNN experiments on MNIST, the FID and MS scores obtained by MD-GAN and FL-GAN are close to equivalent.
In the CNN experiments on CIFAR10, MD-GAN obtains better IS and MS than FL-GAN over this more complex learning task. 

These three experiments show that MD-GAN exploits the advantage of having a single generator to train, that faces multiple discriminators.

\subsubsection{Scalability and the impact of worker to worker communications} We present on Figure \ref{fig:scale} the evolution of the final accuracy score for MD-GAN (after $20,000$ iterations), as a function of the number of workers using the MLP model. Because the  dataset is split over workers, increasing the number of participants reduces the size of local datasets ($|\mathcal{B}_n|=|\mathcal{B}|/N$). 

Two variants of MD-GAN are executed. The first one is the discussed MD-GAN algorithm, and the second one depicts on the dotted curves MD-GAN where no swapping between workers occurs (\ie with respectively $E=1$, and $E=\infty$). The blue curves present the MD-GAN scores when the workload on workers (\ie the number of images to process) remains constant, while the orange curves present a constant workload of the central node.
We note that Figure \ref{fig:scale} also illustrates a varying size of mini-batches $b$ used by workers on the curve with a constant workload on server: the larger $N$ is, the lower $b$ is in consequence, to maintain the same workload on the server.

We note that interesting phenomenons appear at scale after $N=10$; for lower values of $N$ the workers appear to have enough data locally to reach satisfying scores.

The first observation is that considering a constant workload on workers leads to better results. This yet comes at the price of a higher cost on the server (cf Table \ref{tab:complexities} and \ref{tab:com}).

The swapping process between workers leads to better results. We yet observe that, despite the better result in MS, the FID score improvement using swapping is marginal in the case of the constant workload on server setting. This indicates that data available locally to workers is enough, and that their is a marginal gain to await from the diversity brought by swapping discriminators.

\subsubsection{Fault tolerance of MD-GAN facing worker crashes}

In order to assess the tolerance of a MD-GAN learning task facing worker fail-stop crashes (workers' data also disappear from the system when the crash occurs), we conduct the following experiment, presented in Figure \ref{fig:crashes}.
We operate in the same scenario than for experiments in Figure \ref{fig:re1}, and for the best performing MD-GAN setup (with $k=\lfloor \log(N) \rfloor$), but this time we trigger a worker to crash every $I/N$ iterations (appearing as the curve in green). 
Consequence is that at $I=50,000$, all workers have crashed.
For comparison with a baseline, standalone GAN (\ie single server GAN learning) are reploted for two batch sizes ($b \in [10,100]$), and so is the non crashing run on the blue curve with same parametrization.

First observation is that this crash pattern has a no significant impact on the result performance for the MNIST dataset, for both MS and FD metrics. The MLP architecture even exhibits the smallest FD at the end of the experiment. This highlights that for this dataset, the MD-GAN architecture manages to learn fast enough so that crashes, and then the removal of dataset shares, are not a problem performance-wise. 

Both metrics are affected in the case of the CIFAR10 dataset: we observe a divergence due to crashes, and it happens early in the learning phase (around $I=5,000$, corresponding to the first crashed worker). This experiment shows the sensitivity of the learning to early failures, because GANs did not have enough time to accurately approximate the distribution of the data, and then misses the lost data shares for reaching a competitive score. Scores are yet comparable to the standalone baseline up to 8 crashed workers.

We nevertheless note that in the geo-distributed learning frameworks \cite{gaia,geo} that our work is aiming to support, the crashes of several workers will undoubtedly trigger repair mechanisms, in order to cope with diverging learning tendencies.

\subsubsection{Validation on a larger dataset}
\begin{figure}
\centering
\includegraphics[width=3.5in]{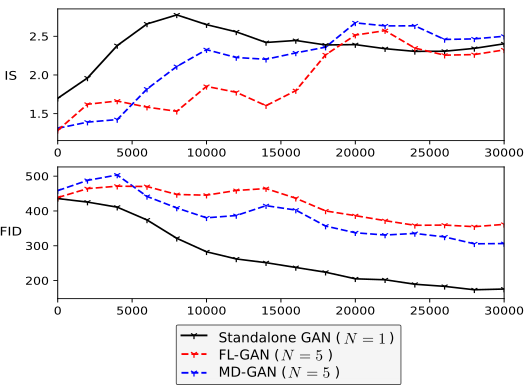}
\caption{Inception scores and Fréchet Inception Distance of the three competitors, on the CelebA dataset.}
\label{fig:celebA}
\end{figure}

In this experiment, we validate the convergence of MD-GAN, and its interest with regards to the standalone and FL-GAN approaches. The goal is to train a GAN over the CelebA dataset \cite{liu2015faceattributes}, which is composed by $200$K images of celebrities ($128\times 128$ pixels). 
We use $10$K images as the test dataset, while the remaining images are distributed equally (\textit{i.i.d.}) over the $N=\{1,5\}$ workers. The GAN architecture is a variant of the one used for the CIFAR10 dataset: \G is composed of one fully-connected layer of $16,384$ neurons and two transposed convolutional layers of respectively $128$ and $3$ kernels of size $5\times5$; \D is composed of six convolutional layers of respectively $16$, $32$, $64$, $128$, $256$ and $512$ kernels of size $3\times3$, and one fully-connected layer of one neuron. The batch size for the standalone GAN and FL-GAN is $b=200$ whereas the batch size of MD-GAN is $b=40$ (corresponding to $200$ images processed to compute one generator update). In this experiment, we use two different settings for the Adam optimizer, leading to better results for each competitor. The standalone GAN and FL-GAN uses a learning rate of $\alpha=0.003$  (resp. $\alpha=0.002$), $\beta_1=0.5$ (resp. $\beta_1=0.5$) and $\beta_2=0.999$ (resp. $\beta_2=0.999$) for the optimizer of \G  (resp. optimizer of \D), whereas MD-GAN uses a learning rate of $\alpha=0.001$  (resp. $\alpha=0.004$), $\beta_1=0.0$ (resp. $\beta_1=0.0$) and $\beta_2=0.9$ (resp. $\beta_2=0.9$) for optimizer of \G (resp. optimizer of \D). The resulting FID and Inception scores during the $30,000$ iterations we considered are reported in Figure \ref{fig:celebA}.
We observe that all IS scores are comparable (MD-GAN is slightly above); yet regarding the FID, MD-GAN (as well as FL-GAN) is distanced by the standalone approach (as this is the case for the CNN experiment on MNIST).

\section{Related Work: Distributing Deep Learning}
\label{relatex}

Distributing the learning of deep neural networks over multiple machines is generally performed with the Parameter Server model proposed by J. Dean et al in \cite{DistBelief_DownpourSGD_2012}. This model was adapted in different works \cite{Rudra_2015,Staleness-awareASGD_2015,ElasticASGD}. The first interest is to speed up the learning in large data-centers \cite{ParameterServer_OSDI2014,Project-Adam_OSDI2014,SyncSGD_2016}. This parameter server model was used for privacy reasons in \cite{PrivacyPreservingDDL_Shokri_2015}. The federated learning is a most accomplished method using the parameter server model with auxiliary workers to reduce communications \cite{FederatedLearning_Comm_2016} or increase the privacy \cite{cryptFed}.


We experimented in a position paper \cite{Hardy:2018:GGP:3286490.3286563} the distribution of the generator function. In this fully decentralized setup where compute nodes exchange their generators and discriminators in a gossip fashion (there are $n$ couples of generator and discriminators, one per worker), the experiment results are favorable to federated learning. We then propose MD-GAN as a solution for a performance gain over federated learning.

Finally, a recent work \cite{GAP} proposes to multiply the number of discriminators and generators in a datacenter location: the authors propose to train several couples of GAN in parallel and to swap generators and discriminators every fix amount of iterations. Durugkar et al. \cite{MGAN} propose a centralized multi-discriminators architecture to improve the discriminator judgment on generated data. In the same way, Hoang et al. \cite{mGGAN} study a centralized multi-generator architecture is proposed to improve the generator capacities and to reduce the so called \textit{mode collapse problem} \cite{principledMethodsGAN}. The works \cite{EnsembleGANs} and \cite{AdaGAN} improve the mixture of generative adversarial models. Wang et al. \cite{EnsembleGANs} use ensemble of GANs trained separately organized as a cascade to build an aggregated model. In the work of Tolstikhin et al. \cite{AdaGAN}, GANs are trained sequentially using boosting strategies to incrementally improve the performance of the final model.
Note that all these works are proposed to improve GAN convergence, but not to distribute the learning (discriminators have access to the whole dataset). Our contribution is a method leveraving multiple adversaries in a distributed setup, and taking the network constraints into account.

\section{Perspectives and conclusion}
\label{conclusion}

Before we conclude, we highlight the salient questions on the way to a widespread distribution of GANs.

\subsubsection{Asynchronous setting}

Instead waiting all $F$ every global iteration, the server may compute a gradient $\Delta w$ and apply it each time it receives a single $F_n$. Fresh batches of data can be generated frequently, so that they can be sent to idle workers. All workers can operate without global synchronization, contrarily to federated learning methods as FL-GAN. In this setting, the waiting time of both workers and the server are reduced drastically. However, because of asynchronous updates, there is no guarantee that the parameters $\bm{w}$ of a worker $n$ at time $t$ (used to generate $X^{(g)}_n$) are the same at time $t+\Delta t$ when it sends its $F_n$ to the server.

In the parameter server model, asynchrony implies inconsistent updates by workers. In practice, the training task nevertheless works well if the learning rate is adapted in consequence \cite{Staleness-awareASGD_2015,Rudra_2015,nca1}.

\subsubsection{The central server communication bottleneck}

The parameter server framework, despite its simplicity, has the obvious drawback of creating a communication bottleneck towards the central server. This has been quantified by several works \cite{ParameterServer_OSDI2014,nca1}, and solutions for traffic reduction between workers and the server have been proposed. 
Methods such as Adacomp \cite{nca1} propose to communicate updates based on gradient staleness, which constitutes a form of data compression. 

In the context of GANs, those methods may be applied on generated data before they are sent to workers, and to the error feedback messages sent by workers to the server. In particular, concerning images data, there are many techniques from their compression (with or without loss of information, see \eg \cite{imgcomp}).

A fine grained combination of techniques for gradient and data object compression would make the parameter server framework more sustainable for GANs and increasingly larger datasets to learn on.
A second direction might be to mix the federated learning approach with ensemble of GANs training independently in cascades (as presented by  Wang et al. \cite{EnsembleGANs}). Federated learning would act as the scheduling mechanism for the parallel ensembles; this would restrict the burden on the server to critical only communications (up-to-date model hosting and dispatching), while most of the training occurs on edge workers, hierarchically.

\subsubsection{Adversaries in generative adversarial networks}

The current deployment setup of GANs in the literature is assuming an adversary-free environment.
In fact, the question of the capacity of basic deep learning mechanisms to embed byzantine fault tolerance has just been recently proposed for distributed gradient descent \cite{bft}.
In addition to the gradient updates in GANs, and more specifically, the learning process is most likely prone to workers having their discriminator lie to the server's generator (by sending erroneous or manipulated feedback). The global convergence, and then the final performance of the learning task will be affected in an unknown proportion. This adversarial setup, and more generally better fault tolerance, are a crucial aspect for future applications in the domain.

\subsubsection{Scaling the number of workers}

We experimented MD-GAN over up to $50$ parallel workers. 
 The current scale at which parallel deep learning is operating is in the order to tens (\eg in Gaia \cite{gaia} or in \cite{geo}) to few hundreds of workers (experiments in TensorFlow \cite{osdi_tf} for instance reach $256$ workers maximum). It is still not well understood what is the bottleneck for reaching larger scales: is the dataset size imposing the scale? Or is this the conflicting asynchronous updates \cite{Staleness-awareASGD_2015} from workers to the server limiting the benefit of scale after a certain threshold? 
We note that federated Learning can be used on a large number of workers (\eg $2,000$ in some works \cite{FederatedLearning_Comm_2016}) by using only a random subset of the available devices at every round. MD-GAN can be adapted in a similar way, with fewer discriminators than workers: because discriminator models are swapped during the learning process, the whole distributed dataset could be leveraged.

Those general questions for deep learning are also applying to the learning of GANs, as they are themselves constituted by couples of deep neural networks. The unknown spot comes from the specificity of GANs, because of the coupling of generators and discriminators; that coupling will most likely play an additional major role in the future algorithms that will be dedicated to push the scalability of GANs to a new standard.

\vspace{0.3cm}
This paper has presented generative adversarial networks in the novel context of parallel computation and of learning over distributed datasets; MD-GAN aims at being leveraged by geo-distributed or edge-device deep learning setups.
We have presented an adaptation of federated learning to the problem of distributing GANs, and shown that it is possible to propose an algorithm (MD-GAN) that removes half the computation complexity from workers by using a discriminator swapping technique, while still achieving better results on the two reviewed datasets. 
GANs are computationaly and communication intensive, specially in the considered data-distributed setup; we believe this work brought a first viable solution to that domain. We hope that raised perspectives will trigger interesting future works for the system and algorithmic support of the nascent field of generative adversarial networks. 

\bibliographystyle{IEEEtran}
\bibliography{main}

\begin{thebibliography}{10}
\providecommand{\url}[1]{#1}
\csname url@samestyle\endcsname
\providecommand{\newblock}{\relax}
\providecommand{\bibinfo}[2]{#2}
\providecommand{\BIBentrySTDinterwordspacing}{\spaceskip=0pt\relax}
\providecommand{\BIBentryALTinterwordstretchfactor}{4}
\providecommand{\BIBentryALTinterwordspacing}{\spaceskip=\fontdimen2\font plus
\BIBentryALTinterwordstretchfactor\fontdimen3\font minus
  \fontdimen4\font\relax}
\providecommand{\BIBforeignlanguage}[2]{{%
\expandafter\ifx\csname l@#1\endcsname\relax
\typeout{** WARNING: IEEEtran.bst: No hyphenation pattern has been}%
\typeout{** loaded for the language `#1'. Using the pattern for}%
\typeout{** the default language instead.}%
\else
\language=\csname l@#1\endcsname
\fi
#2}}
\providecommand{\BIBdecl}{\relax}
\BIBdecl

\bibitem{gan}
I.~J. {Goodfellow}, J.~{Pouget-Abadie}, M.~{Mirza}, B.~{Xu}, D.~{Warde-Farley},
  S.~{Ozair}, A.~{Courville}, and Y.~{Bengio}, ``{Generative Adversarial
  Networks},'' \emph{ArXiv e-prints}, Jun. 2014.

\bibitem{ganText2Img}
S.~{Reed}, Z.~{Akata}, X.~{Yan}, L.~{Logeswaran}, B.~{Schiele}, and H.~{Lee},
  ``{Generative Adversarial Text to Image Synthesis},'' \emph{ArXiv e-prints},
  May 2016.

\bibitem{ganImg2Vid}
C.~{Vondrick}, H.~{Pirsiavash}, and A.~{Torralba}, ``{Generating Videos with
  Scene Dynamics},'' \emph{ArXiv e-prints}, Sep. 2016.

\bibitem{resolution}
C.~{Ledig}, L.~{Theis}, F.~{Huszar}, J.~{Caballero}, A.~{Cunningham},
  A.~{Acosta}, A.~{Aitken}, A.~{Tejani}, J.~{Totz}, Z.~{Wang}, and W.~{Shi},
  ``{Photo-Realistic Single Image Super-Resolution Using a Generative
  Adversarial Network},'' \emph{CVPR}, 2017.

\bibitem{editImg}
G.~{Perarnau}, J.~{van de Weijer}, B.~{Raducanu}, and J.~M. {{\'A}lvarez},
  ``{Invertible Conditional GANs for image editing},'' \emph{ArXiv e-prints},
  Nov. 2016.

\bibitem{DBLP:journals/corr/ChidambaramQ17}
M.~Chidambaram and Y.~Qi, ``Style transfer generative adversarial networks:
  Learning to play chess differently,'' \emph{CoRR}, vol. abs/1702.06762, 2017.

\bibitem{anomaly}
E.~J. Hyunsun~Choi, ``Generative ensembles for robust anomaly detection,''
  \emph{CoRR}, vol. abs/1810.01392v1, 2018.

\bibitem{gaia}
K.~Hsieh, A.~Harlap, N.~Vijaykumar, D.~Konomis, G.~R. Ganger, P.~B. Gibbons,
  and O.~Mutlu, ``Gaia: Geo-distributed machine learning approaching {LAN}
  speeds,'' in \emph{NSDI}, 2017.

\bibitem{geo}
I.~Cano, M.~Weimer, D.~Mahajan, C.~Curino, and G.~M. Fumarola, ``Towards
  geo-distributed machine learning,'' \emph{CoRR}, vol. abs/1603.09035, 2016.

\bibitem{MGAN}
I.~{Durugkar}, I.~{Gemp}, and S.~{Mahadevan}, ``{Generative Multi-Adversarial
  Networks},'' \emph{5th International Conference on Learning Representations
  (ICLR 2017)}, Nov. 2016.

\bibitem{mGGAN}
Q.~{Hoang}, T.~{Dinh Nguyen}, T.~{Le}, and D.~{Phung}, ``{Multi-Generator
  Generative Adversarial Nets},'' \emph{ArXiv e-prints}, Aug. 2017.

\bibitem{ParameterServer_OSDI2014}
M.~Li, D.~G. Andersen, J.~W. Park, A.~J. Smola, A.~Ahmed, V.~Josifovski,
  J.~Long, E.~J. Shekita, and B.-Y. Su, ``Scaling distributed machine learning
  with the parameter server,'' in \emph{OSDI}, 2014.

\bibitem{nca1}
C.~Hardy, E.~{Le Merrer}, and B.~Sericola, ``Distributed deep learning on
  edge-devices: Feasibility via adaptive compression,'' in \emph{NCA}, 2017.

\bibitem{Staleness-awareASGD_2015}
W.~Zhang, S.~Gupta, X.~Lian, and J.~Liu, ``Staleness-aware async-sgd for
  distributed deep learning,'' 2016.

\bibitem{FederatedLearning_premices}
H.~B. McMahan, E.~Moore, D.~Ramage, and B.~A. y~Arcas, ``Federated learning of
  deep networks using model averaging,'' \emph{CoRR}, vol. abs/1602.05629,
  2016.

\bibitem{adam}
D.~P. Kingma and J.~Ba, ``Adam: {A} method for stochastic optimization,''
  \emph{CoRR}, vol. abs/1412.6980, 2014.

\bibitem{principledMethodsGAN}
M.~{Arjovsky} and L.~{Bottou}, ``{Towards Principled Methods for Training
  Generative Adversarial Networks},'' \emph{ArXiv e-prints}, Jan. 2017.

\bibitem{wgan}
M.~{Arjovsky}, S.~{Chintala}, and L.~{Bottou}, ``{Wasserstein GAN},''
  \emph{ArXiv e-prints}, Jan. 2017.

\bibitem{acgan}
A.~Odena, C.~Olah, and J.~Shlens, ``Conditional image synthesis with auxiliary
  classifier {GAN}s,'' in \emph{Proceedings of the 34th International
  Conference on Machine Learning}, ser. Proceedings of Machine Learning
  Research, D.~Precup and Y.~W. Teh, Eds., vol.~70.\hskip 1em plus 0.5em minus
  0.4em\relax International Convention Centre, Sydney, Australia: PMLR, 06--11
  Aug 2017, pp. 2642--2651.

\bibitem{ImprovedGAN}
T.~{Salimans}, I.~{Goodfellow}, W.~{Zaremba}, V.~{Cheung}, A.~{Radford}, and
  X.~{Chen}, ``{Improved Techniques for Training GANs},'' \emph{ArXiv
  e-prints}, Jun. 2016.

\bibitem{dynamo}
G.~DeCandia, D.~Hastorun, M.~Jampani, G.~Kakulapati, A.~Lakshman, A.~Pilchin,
  S.~Sivasubramanian, P.~Vosshall, and W.~Vogels, ``Dynamo: Amazon's highly
  available key-value store,'' in \emph{SOSP}, 2007.

\bibitem{DistBelief_DownpourSGD_2012}
J.~Dean, G.~Corrado, R.~Monga, K.~Chen, M.~Devin, M.~Mao, M.~aurelio Ranzato,
  A.~Senior, P.~Tucker, K.~Yang, Q.~V. Le, and A.~Y. Ng, ``Large scale
  distributed deep networks,'' in \emph{NIPS}, F.~Pereira, C.~J.~C. Burges,
  L.~Bottou, and K.~Q. Weinberger, Eds., 2012.

\bibitem{gossipDL}
M.~{Blot}, D.~{Picard}, M.~{Cord}, and N.~{Thome}, ``{Gossip training for deep
  learning},'' \emph{ArXiv e-prints}, Nov. 2016.

\bibitem{Hardy:2018:GGP:3286490.3286563}
C.~Hardy, E.~{Le Merrer}, and B.~Sericola, ``Gossiping {GAN}s,'' in
  \emph{Proceedings of the Second Workshop on Distributed Infrastructures for
  Deep Learning}, ser. DIDL '18, 2018.

\bibitem{DBLP:journals/corr/WangZW16}
Y.~Wang, L.~Zhang, and J.~van~de Weijer, ``Ensembles of generative adversarial
  networks,'' in \emph{NIPS 2016 Workshop on Adversarial Training}, 2016.

\bibitem{NIPS2017_7126}
\BIBentryALTinterwordspacing
I.~O. Tolstikhin, S.~Gelly, O.~Bousquet, C.-J. SIMON-GABRIEL, and
  B.~Sch\"{o}lkopf, ``Adagan: Boosting generative models,'' in \emph{Advances
  in Neural Information Processing Systems 30}, I.~Guyon, U.~V. Luxburg,
  S.~Bengio, H.~Wallach, R.~Fergus, S.~Vishwanathan, and R.~Garnett, Eds.\hskip
  1em plus 0.5em minus 0.4em\relax Curran Associates, Inc., 2017, pp.
  5424--5433. [Online]. Available:
  \url{http://papers.nips.cc/paper/7126-adagan-boosting-generative-models.pdf}
\BIBentrySTDinterwordspacing

\bibitem{FederatedLearning_Comm_2016}
J.~{Kone{\v c}n{\'y}}, H.~{Brendan McMahan}, F.~X. {Yu}, P.~{Richt{\'a}rik},
  A.~{Theertha Suresh}, and D.~{Bacon}, ``{Federated Learning: Strategies for
  Improving Communication Efficiency},'' \emph{CoRR}, vol. abs/1610.05492, Oct.
  2016.

\bibitem{Hogwild_NIPS2011}
B.~Recht, C.~Re, S.~Wright, and F.~Niu, ``Hogwild: A lock-free approach to
  parallelizing stochastic gradient descent,'' in \emph{NIPS}, 2011.

\bibitem{AsyncSGD_NIPS2015}
\BIBentryALTinterwordspacing
X.~Lian, Y.~Huang, Y.~Li, and J.~Liu, ``Asynchronous parallel stochastic
  gradient for nonconvex optimization,'' in \emph{NIPS}, C.~Cortes, N.~D.
  Lawrence, D.~D. Lee, M.~Sugiyama, and R.~Garnett, Eds.\hskip 1em plus 0.5em
  minus 0.4em\relax Curran Associates, Inc., 2015. [Online]. Available:
  \url{http://papers.nips.cc/paper/5751-asynchronous-parallel-stochastic-gradient-for-nonconvex-optimization.pdf}
\BIBentrySTDinterwordspacing

\bibitem{SSGD}
J.~{Chen}, X.~{Pan}, R.~{Monga}, S.~{Bengio}, and R.~{Jozefowicz},
  ``{Revisiting Distributed Synchronous SGD},'' \emph{arXiv e-prints}, p.
  arXiv:1604.00981, Apr. 2016.

\bibitem{Rudra_2015}
S.~Gupta, W.~Zhang, and F.~Wang, ``Model accuracy and runtime tradeoff in
  distributed deep learning: A systematic study,'' in \emph{ICDM}, Dec 2016.

\bibitem{GoogleNet_2015}
C.~Szegedy, W.~Liu, Y.~Jia, P.~Sermanet, S.~Reed, D.~Anguelov, D.~Erhan,
  V.~Vanhoucke, and A.~Rabinovich, ``Going deeper with convolutions,'' in
  \emph{CVPR}, 2015.

\bibitem{lecun1998mnist}
\BIBentryALTinterwordspacing
Y.~LeCun, C.~Cortes, and C.~J. Burges, ``The mnist database of handwritten
  digits,'' \url{http://yann.lecun.com/exdb/mnist}, 1998. [Online]. Available:
  \url{http://yann.lecun.com/exdb/mnist/}
\BIBentrySTDinterwordspacing

\bibitem{cifar10}
A.~Krizhevsky, ``Learning multiple layers of features from tiny images,'' 2009.

\bibitem{FID}
M.~{Heusel}, H.~{Ramsauer}, T.~{Unterthiner}, B.~{Nessler}, and
  S.~{Hochreiter}, ``{GANs Trained by a Two Time-Scale Update Rule Converge to
  a Local Nash Equilibrium},'' \emph{ArXiv e-prints}, Jun. 2017.

\bibitem{liu2015faceattributes}
Z.~Liu, P.~Luo, X.~Wang, and X.~Tang, ``Deep learning face attributes in the
  wild,'' in \emph{Proceedings of International Conference on Computer Vision
  (ICCV)}, 2015.

\bibitem{ElasticASGD}
S.~Zhang, A.~E. Choromanska, and Y.~LeCun, ``Deep learning with elastic
  averaging {SGD},'' in \emph{NIPS}, C.~Cortes, N.~D. Lawrence, D.~D. Lee,
  M.~Sugiyama, and R.~Garnett, Eds., 2015.

\bibitem{Project-Adam_OSDI2014}
T.~Chilimbi, Y.~Suzue, J.~Apacible, and K.~Kalyanaraman, ``Project adam:
  Building an efficient and scalable deep learning training system,'' in
  \emph{OSDI}, 2014.

\bibitem{SyncSGD_2016}
J.~Chen, R.~Monga, S.~Bengio, and R.~Jozefowicz, ``Revisiting distributed
  synchronous {SGD},'' 2016.

\bibitem{PrivacyPreservingDDL_Shokri_2015}
R.~Shokri and V.~Shmatikov, ``Privacy-preserving deep learning,'' in
  \emph{CCS}, 2015.

\bibitem{cryptFed}
K.~Bonawitz, V.~Ivanov, B.~Kreuter, A.~Marcedone, H.~B. McMahan, S.~Patel,
  D.~Ramage, A.~Segal, and K.~Seth, ``Practical secure aggregation for privacy
  preserving machine learning,'' Cryptology ePrint Archive, Report 2017/281,
  2017.

\bibitem{GAP}
D.~{Jiwoong Im}, H.~{Ma}, C.~{Dongjoo Kim}, and G.~{Taylor}, ``{Generative
  Adversarial Parallelization},'' \emph{ArXiv e-prints}, Dec. 2016.

\bibitem{EnsembleGANs}
Y.~{Wang}, L.~{Zhang}, and J.~{van de Weijer}, ``{Ensembles of Generative
  Adversarial Networks},'' \emph{arXiv e-prints}, p. arXiv:1612.00991, Dec.
  2016.

\bibitem{AdaGAN}
I.~{Tolstikhin}, S.~{Gelly}, O.~{Bousquet}, C.-J. {Simon-Gabriel}, and
  B.~{Sch{\"o}lkopf}, ``{AdaGAN: Boosting Generative Models},'' \emph{arXiv
  e-prints}, p. arXiv:1701.02386, Jan. 2017.

\bibitem{imgcomp}
M.~J. Weinberger, G.~Seroussi, and G.~Sapiro, ``The loco-i lossless image
  compression algorithm: principles and standardization into jpeg-ls,''
  \emph{IEEE Transactions on Image Processing}, vol.~9, no.~8, pp. 1309--1324,
  Aug 2000.

\bibitem{bft}
P.~Blanchard, E.~M.~E. Mhamdi, R.~Guerraoui, and J.~Stainer, ``Machine learning
  with adversaries: Byzantine tolerant gradient descent,'' in \emph{NIPS},
  2017.

\bibitem{osdi_tf}
M.~Abadi, P.~Barham, J.~Chen, Z.~Chen, A.~Davis, J.~Dean, M.~Devin,
  S.~Ghemawat, G.~Irving, M.~Isard, M.~Kudlur, J.~Levenberg, R.~Monga,
  S.~Moore, D.~G. Murray, B.~Steiner, P.~Tucker, V.~Vasudevan, P.~Warden,
  M.~Wicke, Y.~Yu, and X.~Zheng, ``Tensorflow: A system for large-scale machine
  learning,'' in \emph{OSDI}, 2016.

\end{thebibliography}

\end{document}